# Numerical Energy Analysis of In-wheel Motor Driven Autonomous Electric Vehicles

Kang Shen[a], Fan Yang[a], Xinyou Ke[a], Cheng Zhang[b], Chris Yuan[a,*]

[a]Department of Mechanical and Aerospace Engineering, Case Western Reserve University, Cleveland, Ohio 44106, USA

[b]School of Mechanical Engineering, Nantong University, Nantong, Jiangsu 226019, China

*Corresponding author: chris.yuan@case.edu

## Abstract

Autonomous electric vehicles are being widely studied nowadays as the future technology of ground transportation, while the autonomous electric vehicles based on conventional powertrain system limit their energy and power transmission efficiencies and may hinder their broad applications in future. Here we report a study on the energy consumption and efficiency improvement of a mid-size autonomous electric vehicle driven by in-wheel motors, through the development of a numerical energy model, validated with the actual driving data and implemented in a case study. The energy analysis was conducted under three driving conditions: flat road, upslope, and downslope driving to examine the energy consumption, with the energy-saving potential of the in-wheel-motor driven powertrain system systematically explored and discussed. Considering the energy recovery from the regenerative braking, energy consumption and regenerated energy were calculated in specific driving cycles based on vehicle dynamics and autonomous driving patterns. A case study was conducted using the baseline electric vehicle driving data in West Los Angeles. It was found that an in-wheel motor driven autonomous electric vehicle can save up to 17.5% of energy compared with a conventional electric vehicle during the slope driving. Using the efficiency maps of a commercial in-wheel motor, the numerical energy model and validated results obtained from this study are in line with actual situations, and can be used to support sustainable development of more energy-efficient autonomous electric vehicles in the future.

***Keywords*:** Autonomous electric vehicle; in-wheel motor; energy efficiency; energy analysis; predictive modeling

# Nomenclature

| Symbol | Units | Definitions |
|---|---|---|
| $a_x$ | $m \cdot s^{-2}$ | vehicle acceleration |
| $A$ | $m^2$ | frontal area |
| $C_D$ | - | drag coefficient |
| $D$ | $m$ | inner diameter of stator |
| $D_{cycle}$ | $km$ | driving distance of each driving cycle |
| $E$ | $Wh$ | energy consumption per km |
| $E_{reg}$ | $Wh$ | regenerated energy per km |
| $E_{unit}$ | $Wh$ | unit energy consumption for each driving cycle |
| $f_R$ | - | road friction coefficient |
| $F_g$ | $N$ | slope resistance |
| $F_a$ | $N$ | acceleration resistance |
| $F_r$ | $N$ | rolling resistance |
| $F_d$ | $N$ | aerodynamic resistance |
| $F_{dem}$ | $N$ | demanding traction force |
| $K_{sp}$ | $A \cdot m^{-1}$ | fundamental linear current density |
| $K_{fill}$ | - | filling factor |
| $L$ | $m$ | length of stator lamination |
| $L_{ew}$ | $m$ | end-winding length |
| $m_c$ | $kg$ | mass of the cargo |
| $m_v^*$ | $kg$ | baseline EV mass |
| $m_v$ | $kg$ | IWM-AEV mass |
| $n$ | $rpm$ | rotating speed of wheels |
| $p$ | - | number of pole pairs |
| $P_{cu}$ | $W$ | stator copper loss |
| $P_e$ | $W$ | electromagnetic power |
| $P_{Fe}$ | $W$ | stator core loss |
| $P_{mag}$ | $W$ | rotor magnet loss |
| $P_{recovery}$ | $kW$ | energy recovery of the traction system |
| $P_{SSCM}$ | $kW$ | power demand of the SSCM system |
| $P_{traction}$ | $kW$ | power demand of the traction system |
| $r_d$ | $m$ | tire rolling radius |
| $T_{dem,1}$ | $N \cdot m$ | upslope and falt road demanding traction torque |
| $T_{dem,2}$ | $N \cdot m$ | torque demand to balance the electric vehicle during braking |
| $v$ | $km \cdot h^{-1}$ | vehicle driving speed |
| $Vol_t$ | - | the volume of stator tooth |
| $Vol_j$ | - | the volume of stator yoke |
| $\lambda$ | - | demanding energy factor |
| $\lambda_2$ | - | regenerating energy factor |
| $\theta$ | ° | slope angle |
| $\delta_i$ | - | vehicle rotating mass conversion factor |
| $\eta_b$ | - | IWM braking efficiency |
| $\eta_c$ | - | battery charging efficiency |
| $\eta_d$ | - | battery discharging efficiency |
| $\eta_i^*$ | - | baseline EV inverter efficiency |
| $\eta_i$ | - | IWM-AEV inverter efficiency |
| $\eta_m$ | - | IWM motoring efficiency |
| $\eta_{recover}$ | - | IWM braking energy recovery rate |
| $\eta_t^*$ | - | baseline EV transmission efficiency |
| $\eta_t$ | - | IWM-AEV transmission efficiency |
| $\rho_a$ | $kg \cdot m^{-3}$ | air density |
| $\rho_{Cu}$ | $\Omega \cdot m$ | copper resistivity |
| $\rho_s$ | $kg \cdot m^{-3}$ | mass density of lamination |

1. **Introduction**

Autonomous electric vehicles (AEVs) have recently attracted enormous research interests as next-generation transportation technologies for reducing energy consumption and greenhouse gas (GHG) emissions from the transportation sector. AEV is an electric vehicle (EV) driven by an autonomous control system. In general, an AEV has three key components: energy source (battery), actuators (powertrain), and the self-driving sensing and computing module (SSCM). The battery provides power to all the on-board energy-consuming devices. The on-board computer processes the real-time signal inputs from various sensors and then generates commands of the driving routes and behaviors to be passed to actuators, which will follow the commands from SSCM to control vehicle motions. For an AEV, most of the battery energy is consumed by the traction system to overcome the driving resistances. Despite the largest share of the energy going to the wheels, there are also many other energy consumers in the AEV. For instance, SSCM needs the power to make the autonomous driving system work; the auxiliary devices, such as heating, ventilation, and air conditioning system (HVAC), entertainment system, and lighting system in the AEV also consume some energy [1]. Additionally, part of the battery energy is lost due to the energy transmission efficiency loss in the inverter, electric motor and mechanical transmission system, such as gearbox and driving shaft [2]. In general, an AEV has advantages over a conventional electric vehicle in saving energy since the driver is eliminated to reduce the mass of motion and the auxiliary devices can be turned off when no passengers are inside. Besides, the dispatch of AEVs can be trip-specific under optimal control and can effectively shorten passenger searching time and driving distance [3].

Since the AEV has advantages over the conventional EV on its intelligence and connectivity, most recent research on AEV energy-saving methods either discussing the intelligent control techniques to optimize the energy efficiency or studying the fleet operation strategy to save the collective energy consumption. For example, Al-Jazaeri et al. proposed an energy-saving fuzzy logic control (FLC) method for AEV, and they concluded that the speed controller can save up to 35% more energy than the proportional integral derivative (PID) controller without compromising performance [4]. Tony et al. designed a smart charging framework for the AEV fleets to reduce the energy waste by 18.2% based on the vehicle grid interaction and validated it by a case study in the Puget Sound region [5]. In addition to these aforementioned aspects, new powertrain layouts designs are being proposed to improve the energy efficiency of the AEV. In recent years, an innovative drive technology called in-wheel motor (IWM) has been investigated as an alternative technology to improve the transmission efficiency of the powertrain system. IWM is an electric motor fixed on the wheel hub, which integrates the powering, transmission and braking systems inside, and directly drives the wheel [6]. Prior to the IWM layout, most EVs used the front-engine & front-wheel-drive (FF) layout and the front-engine & rear-wheel-drive layout (FR layout) [7], and past research on AEV powertrain energy efficiency was mainly based on those conventional layouts [8-10].

IWM technology has great potential in reducing EV's energy consumption according to recent research. First, the IWM layout has a simple structure and high transmission efficiency. The transmission efficiency for the in-wheel motor layout can be improved by 8%~15% compared with the conventional FF layout [11]. Without using transmission parts, the vehicle will be lighter and has more space to support a larger battery pack, which can increase the driving range of the vehicle. In-wheel motors also have a high braking energy recovery rate than the other electric motors, because they are directly connected to the wheels [12]. Second, several energy-saving controllers have been designed for IWMs to improve both vehicle stability and energy efficiency. For example,

Kavuma et al. designed a continuous steering stability controller based on an energy-saving torque distribution algorithm for an EV with four independent-drive IWMs [13]. The controller can improve vehicle steering stability and reduce energy consumption by 23.7% compared with the conventional servo and ordinary continuous controllers. Wu et al. proposed a model predictive control (MPC) method for the four-wheel-driven EV to reduce the electronic control unit (ECU) calculation time and energy consumption based on the road slope information and found that the energy can be saved by 1.27% for the real road condition [14]. Toshihiro et al. designed a range extension system for a four IWM driven EV by controlling the yaw rate and side-slip angle, and reported that it can decrease the energy consumption by 13.4% compared with the conventional driving pattern [15]. Thirdly, for IWM-driven vehicles, it is easy to detect real-time driving and braking forces between tires and road surfaces, and the complex torque allocation plans can be applied to the IWM layout to improve the technical performance and energy efficiency [16, 17]. Li et al. proposed a torque distribution control method for EVs with four IWMs under urban driving conditions based on the motor efficiency map [18]. Compared with a working condition that has an equal drive torque and a fixed-ratio regenerative braking force distribution for four wheels, this energy-saving control method can reduce the overall energy consumption by 7.4%. Jiang et al. designed an optimal torque energy-saving allocation method for a small-size IWM-EV, which can save 117 and 426 kJ per km compared with four-wheel torque equal distribution and rear axle drive. In their study, a prototype IWM-EV was built and a bench test was carried out to verify the validity of the method [19].

The IWM technology is expected to become the dominant powertrain structure in the future due to its better performance in mobility and reliability, simpler transmission system, more precise and independent torque control [20], and more compact chassis integration [21]. However, after our rigorous literature survey, we found that currently no study has been conducted on the IWM driven AEV to analyze its energy consumption pattern and potential in improving the energy efficiency of AEV systems. The aforementioned studies discussed the feasibility of improving IWM-EV's energy efficiency by optimizing the torque distribution among each wheel and the road-wheel interaction based on the IWM model built in the software, such as CarSim [13-18]. However, these reported work has not considered the mass increase of the IWMs and the SSCM, and their energy consumptions and interactions with the powertrain system have not been analyzed and compared with those of conventional EVs either. Besides, the energy analysis and energy-saving design methods on reported IWM-EVs work only focused on particular circumstances, such as the double-lane-change maneuver in CarSim and urban driving cycles without slopes. The possibility of combining the technologies of IWM and AEV has not been discussed in the literature. Additionally, most control methods reported are on light vehicle simulations using the IWMs with a maximum torque less than 600 N·m, which does not meet the actual demand of a vehicle. Thus, there is currently a gap of knowledge between theoretical research and practical application of the in-wheel motor driven autonomous electric vehicle (IWM-AEV) system.

In this paper, we present a numerical analysis of driving energy for IWM-AEV. Using a mid-size commercial EV as the baseline, an AEV model has been configured with commercial IWMs and SSCM systems. Two energy-determining factors, demanding energy factor and regenerating energy factor, are proposed based on the adjusted IWM characteristic contours. Energy analysis on the IWM-AEV with various slope angles and speeds were conducted to analyze the energy consumption under three driving conditions: flat road, upslope and downslope. A case study in West Los Angeles with real travel data and the actual driving data from Tesla Roadster were used to demonstrate and validate the analysis results. The designed IWM-AEV can provide a maximum torque of 2500 N·m and a maximum driving speed of 192 km h$^{-1}$, which makes this study close to the actual vehicle applications and the results can be useful to support the design of energy-efficient IWM-AEV system in future.

## 2. Configuration of the in-wheel motor autonomous electric vehicle

The IWM-AEV configuration is modified from a mid-size commercial EV with a conventional FF layout (front-engine, front-wheel-drive layout) by replacing and rearranging the powertrain components as shown in Figure 1. An 80 kW power and 280 N·m torque AC (alternating current) synchronous electric motor is replaced by two in-wheel motors, each of which can provide 64 kW power and 500 N·m torque [34, 35]. The transmission axle and gearbox are removed, with an SSCM added. Taking the Ford Fusion autonomous vehicle test version as the benchmark [22], components in the SSCM system for the proposed IWM-AEV are listed in Table 1.

Table 1: Parameters of the SSCM system

| Component | Model | Power (W) | Mass (kg) | Number | Reference |
|---|---|---|---|---|---|
| LIDAR | Velodyne VLP-16 | 8 | 0.83 | 2 | [23] |
| Radar | Bosch LRR4 | 4.5 | 0.24 | 2 | [24] |
| Camera | Pt. Gray Dragonfly2 | 2.1 | 0.045 | 7 | [25] |
| Sonar | Bosch Ultrasonic | 0.052 | 0.02 | 8 | [26] |
| GPS | NovAtel PwrPak7 | 1.8 | 0.51 | 1 | [27] |
| V2X wireless communication module | Cohda MK5 module | 2.1 | 0.01 | 1 | [28] |
| Computer | Nvidia Drive PX2 | 98 | 5.075 | 2 | [29, 30] |
| Wire harness and case | / | / | 5.7 | / | [31] |

Red dash-dot flow lines in Figure 1 show the control relationship among the components in the driveline: the battery management system (BMS), which manages the lithium-ion battery pack for power inputs and outputs, and the electronic control unit (ECU), which controls two IWMs. Black solid flow lines in Figure 1 show the energy flow in the traction system: the electricity flows from the car charger to the wheels via the battery, the inverter and the IWMs. The energy efficiency of the IWM-AEV was modeled based on the Argonne National Laboratory's experimental test of delivered energy from the EVSE (Electric Vehicle Supply Equipment) to wheels on a mid-size EV [32]. Table 2 lists the main parameters used for the IWM-AEV configuration in this study. Based on the literature [33], weights of motors and transmission parts were set as 72 kg for an 80 kW motor and 80 kg for the transmission parts. Each IWM weighs 31 kg. Although the IWM-AEV system becomes lighter after removing the transmission parts, an SSCM is needed for autonomous driving, and it weighs 19 kg and consumes 240 W power. Based on the data provided by manufacturers [34], the total efficiency of transmission parts and the motor controller was calculated to be 91%.

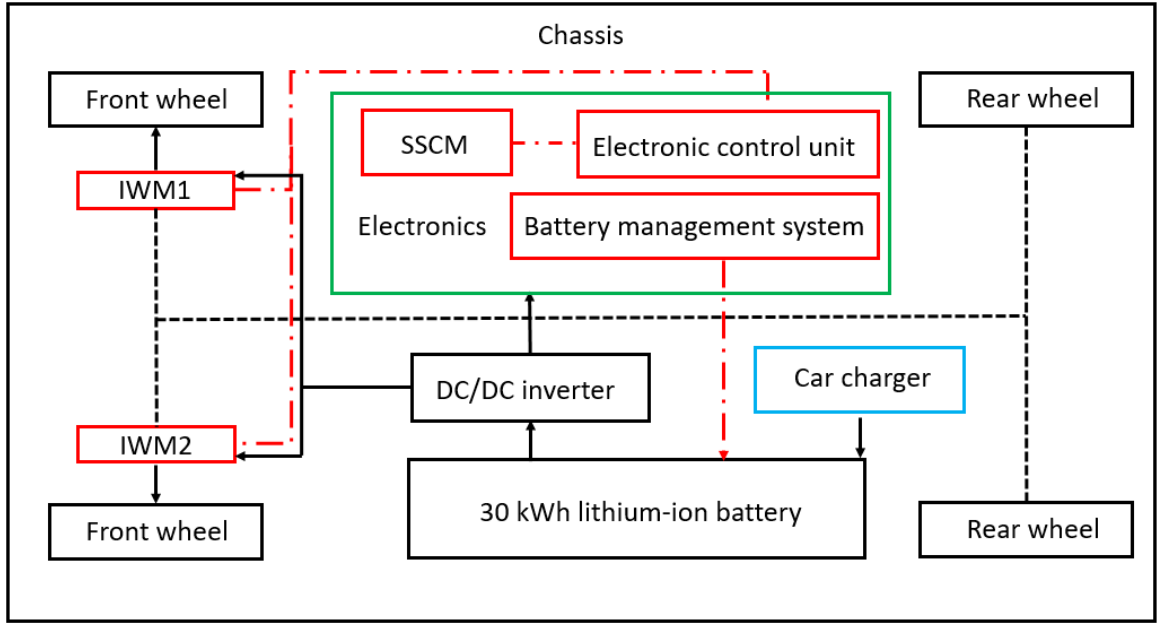

Figure 1: Overall structure of a mid-size IWM-AEV configured in this study.

Table 2: Main parameters used in this study.

| Main parameters | Value | Units | References |
| --- | --- | --- | --- |
| Baseline EV mass ($m_v^*$) | 1481 | $kg$ | [32] |
| IWM-AEV mass ($m_v$) | 1436 | $kg$ | [35] |
| Inertia of rotating parts ($\delta_i * m_v$) | 148 | $kg$ | [36] |
| Frontal area ($A$) | 2.7435 | $m^2$ | [33] |
| Tire rolling radius ($r_d$) | 0.31595 | $m$ | [32] |
| Drag coefficient ($C_D$) | 0.29 | / | [32] |
| Battery charging efficiency ($\eta_c$) | 86.7% | / | [37] |
| Battery discharging efficiency ($\eta_d$) | 88.5% | / | [37] |
| Baseline EV transmission efficiency ($\eta_t^*$) | 93% | / | [32] |
| IWM-AEV transmission efficiency ($\eta_t$) | 100% | / | [12] |

| IWM braking energy recovery rate ($\eta_{recover}$) | 85% | / | [34] |
| Baseline EV inverter efficiency ($\eta_i^*$) | 95% | / | [35] |
| IWM-AEV inverter efficiency ($\eta_i$) | 97.4% | / | [38] |

## 3. Numerical energy analysis of the IWM-AEV driving

*3.1 Driving conditions and strategies for energy analysis*

In this energy analysis, three driving conditions were considered: flat road, upslope and downslope driving, being analyzed in two scenarios. The first scenario covers both upslope and flat road driving conditions as both rely on traction power to move the vehicle. The second scenario considers downslope driving, which requires braking and regenerates energy during driving.

The energy consumed in the driving of a vehicle is mainly governed by various resistances to overcome during moving of the vehicle. In general, the resistance force of the vehicle consists of four parts: the slope resistance ($F_g$), the acceleration resistance ($F_a$), the rolling resistance ($F_r$), and the aerodynamic resistance ($F_d$) [39]. Before carrying out the energy analysis under the downslope condition, the relationships between the slope resistance, and the sum of rolling resistance and aerodynamic resistance are benchmarked. If $F_g > (F_r+F_d)$, the electric motor will work in a braking status since the slope is steep enough to require braking. If $F_g = (F_r+F_d)$, the vehicle resistances are balanced and no external forces are needed. Vice versa, if $F_g < (F_r+F_d)$, the electric motor will work in a motoring status, which is similar to driving the car under the upslope condition. Figure 2 shows the AEV driving logic for energy analysis. The demanding traction force ($F_{dem}$) can be calculated based on the vehicle dynamics. Considering constraints of the IWM efficiency data, the speed points with high energy efficiency can be calculated for both energy consumption and energy regeneration curves. When the IWM-AEV works under the upslope, flat road and downslope motoring conditions, it should operate at the lowest energy consumption speed points. When the IWM-AEV works under the downslope braking condition, it recovers energy from the regenerative braking and should operate at the speed points with the highest energy regeneration. Additionally, if vehicle resistances are balanced, the IWM-AEV can drive down the slope without any energy consumption.

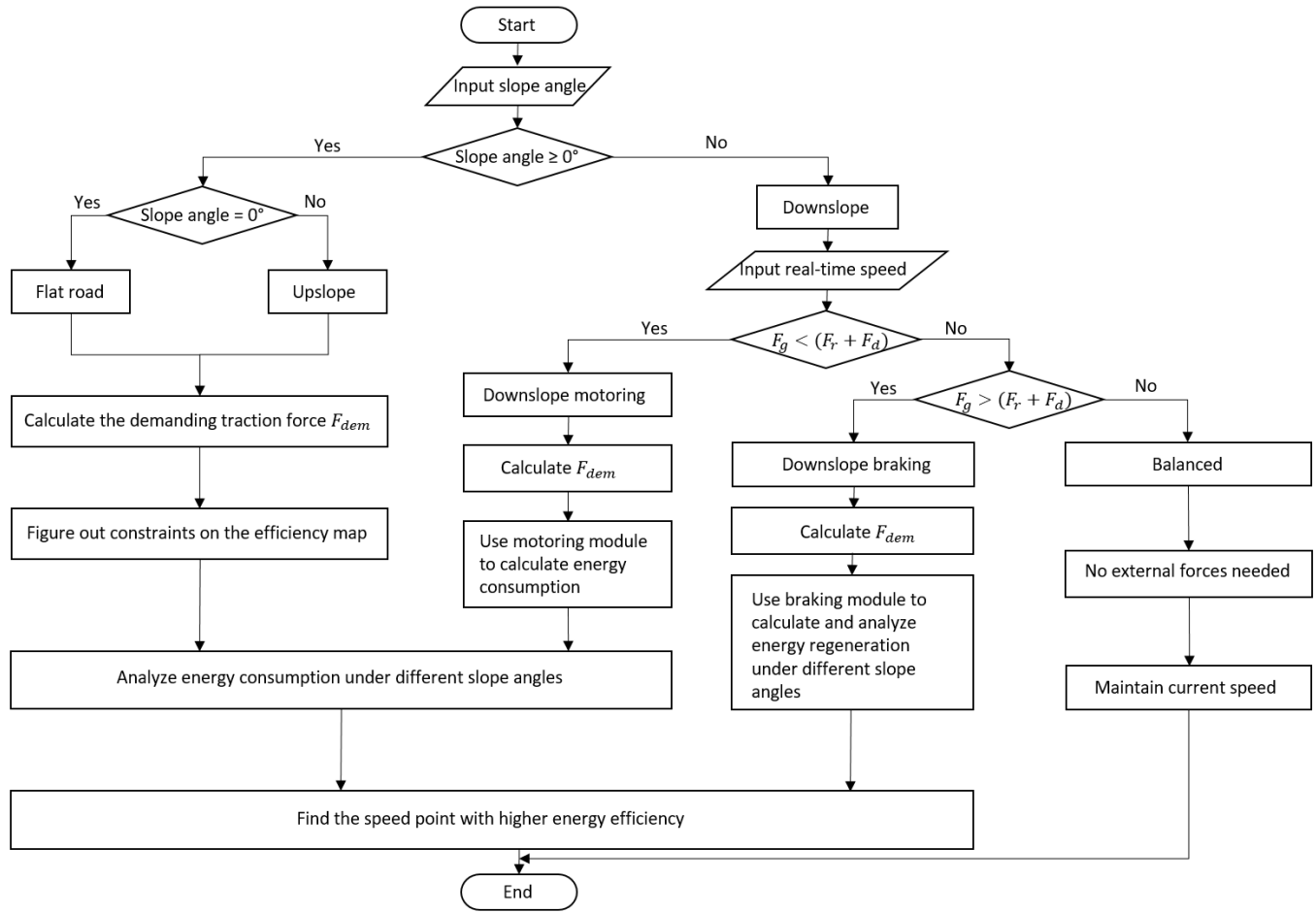

Figure 2: Logic of the energy-saving control for the IWM-AEV driving.

*3.2 IWM efficiency*

The motoring and braking efficiency data were extracted from the official efficiency maps of the commercial IWM used in the aforementioned IWM-AEV design [40], and is plotted into 3D contours as shown in Figure 3 (a) and (b). Both the motoring and braking efficiencies are determined by the torque and rotating speed during the driving. The IWM efficiency varies from 0% to 91% while the rotating speed and torque range from 0 to 1600 rpm (from 0 km h$^{-1}$ to 192 km h$^{-1}$ for the IWM-AEV) and 0 to 1250 N·m, respectively. A small torque or a low speed will lead to low efficiency. High-efficiency areas locate in the middle of the torque and speed range. However, the IWM motoring efficiency map cannot be directly applied to the IWM-AEV energy analysis, because the motor has not been tuned and optimized to make the vehicle operate in high-efficiency areas. It can be seen that over half the area of the motoring efficiency map is lower than 80%. It cannot reflect the actual efficiency of a commercial electric vehicle, compared with an average efficiency of 89% for Nissan Leaf [41] and a maximum efficiency of 97% for Tesla Model 3 [42]. To make the energy analysis more in line with the actual situation, an adjusted motoring efficiency map (see Figure 3 (c)) is produced for our energy analysis based on a detailed IWM model for EV proposed by Z. Li et al [43]. Their work optimized the IWM efficiency for a maximum reduction of the potential energy loss, including the stator copper loss ($P_{Cu}$), the stator core loss ($P_{Fe}$) and the rotor magnet loss ($P_{mag}$). The IWM motoring efficiency can be calculated as follows [43]

$$\eta_m = 1 - \frac{P_{Cu}+P_{Fe}+P_{mag}}{P_e} \qquad (1)$$

$$P_{Cu} = \frac{\frac{\rho_{Cu}}{K_{fill}}(L+L_{ew})(\frac{K_{sp}\pi D_o}{\sqrt{2}k_w})^2}{\frac{\pi}{4}\left(D_o^2-\left(D_i+\frac{B_g\pi D_o}{2B_j pK_{Fe}}\right)^2\right)-\frac{B_g\pi D_o}{B_t}(\frac{D_o-D_i}{2}-\frac{B_g\pi D_o}{4B_j pK_{Fe}})} \qquad (2)$$

$$P_{Fe} = \rho_s(p_{Fe}(B_t)Vol_t + p_{Fe}(B_j)Vol_j) \tag{3}$$

$$P_{mag} = 2pL \sum_{n=1}^{\infty}(p_{cn} + p_{an}) \tag{4}$$

Where, $P_e$ is the electromagnetic power, $\rho_{Cu}$ is the copper resistivity, $K_{fill}$ is the filling factor, $L$ is the length of stator lamination, $L_{ew}$ is the end-winding length, $p$ is the number of pole pairs, $D$ is the inner diameter of stator, $K_{sp}$ is the fundamental linear current density, $\rho_s$ is the mass density of lamination, $Vol_t$ and $Vol_j$ are the volume of stator tooth and yoke, respectively.

From their results, the IWM's maximum motoring efficiency is 94.5% and over 90% of the map area has motoring efficiency higher than 60%. As shown in Figure 3(c), the adjusted motoring efficiency map keeps the contour pattern of the commercial IWM (Figure 3(a)) and falls in the range of reported IWM efficiency in literature between 55% and 94.5% [43]. Here we used the adjusted motoring efficiency in our energy analysis of the IWM-AEV operations.

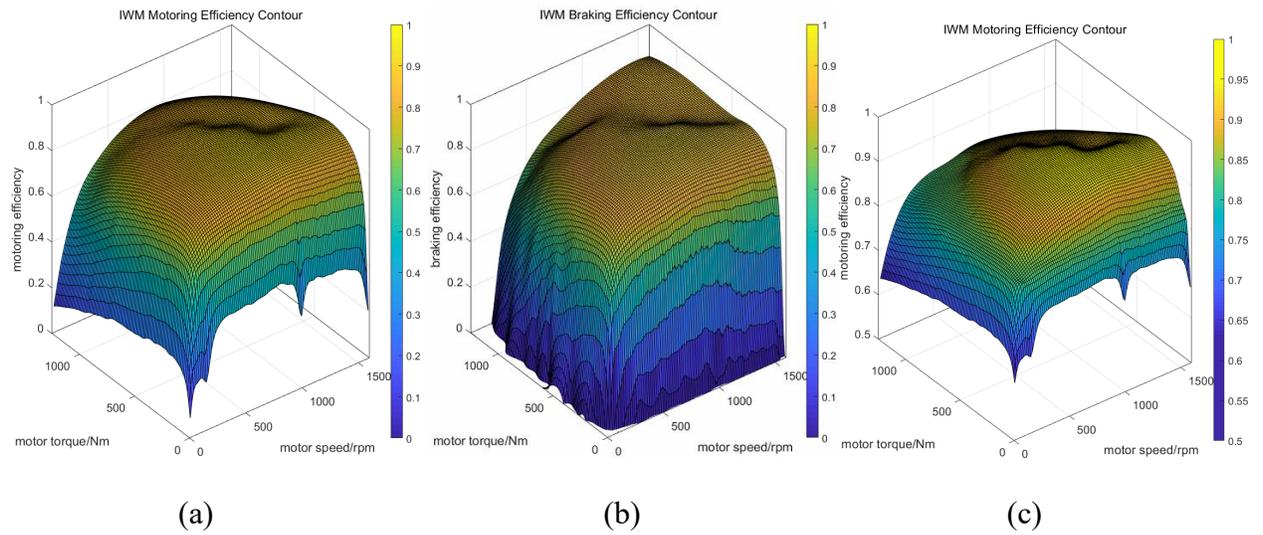

(a)　　　　　　　　　　(b)　　　　　　　　　　(c)

Figure 3: 3D Contour of motor efficiency: (a) motoring efficiency (b) braking efficiency, data extracted from [40]; (c) adjusted motoring efficiency.

*3.3 Upslope and flat road driving*

With the configured IWM motoring efficiency, the energy consumption of the AEVs can be numerically determined for various driving conditions. Considering the flat road and upslope driving in the same scenario, the energy transfer process in the autonomous electric vehicle's powertrain system from the battery to wheels can be modeled as [44]

$$E = E_{traction} + E_{SSCM}$$
$$= P_{traction}t + P_{SSCM}t = \frac{T_{dem,1}*n}{9550*\eta_d\eta_i\eta_m\eta_t} * \frac{1}{v} + P_{SSCM}t$$
$$= \frac{1}{\eta_d\eta_i\eta_m\eta_t}\left(\int F_d v dt + \int F_r v dt + \int F_g v dt + \int_{a>0} F_a v dt\right) - \phi\eta_r \int_{a>0} F_a v dt + P_{SSCM}t \quad (5)$$

Where, $E$ (Wh) is the energy consumption per km, $P_{traction}$ (kW) is the power demand of the traction system, $P_{SSCM}$ (kW) is the power demand of the SSCM system, $T_{dem,1}$ (N·m) is the demanding traction torque, $n$ (rpm) is the rotating speed of wheels, $v$ (km h$^{-1}$) is the driving speed, $\eta_m$ is the IWM motoring efficiency, $\eta_d$ is the battery discharging efficiency, $\eta_t$ is the transmission efficiency from IWM to wheels and $\eta_i$ is the inverter efficiency. The relationship among rotating speed, driving speed and tire rolling radius ($r_d$) gives

$$n = \frac{v}{(2\pi r_d)} * \left(\frac{1000m}{km} * \frac{1h}{60min}\right) = \frac{1}{\pi*r_d*0.12} * v \quad (6)$$

In this analysis, a demanding energy factor ($\lambda=\frac{T_{dem,1}}{\eta_m}$) is defined, and the power demand from the traction systems can be written as:

$$P_{traction} = \frac{30}{\pi * r_d * 3.6 * 9550 * \eta_d * \eta_t * \eta_i} * \lambda \tag{7}$$

Where, the value of the parameters used in Eq. (7) is summarized in Table 2. The power demand of the traction system is a linear function of the demanding energy factor. As a result, the traction energy consumption per km of the autonomous electric vehicle under the upslope driving condition is linearly correlated with the demanding energy factor, which needs to be further analyzed to examine the relationship among energy consumption, slope angle and driving speed.

Under the upslope driving condition, the slope angle and the length of the slope, and the initial speed are known. Decision variables, objective functions and constraints can be figured out before carrying out the energy analysis. For the upslope energy-saving control, the decision variable is defined as the vehicle driving speed $v$, and the objective function is the demanding energy factor ($\lambda$), which is given as

$$\lambda = \frac{T_{dem}}{\eta_m} = \frac{kv^2+b}{\eta_m} \tag{8}$$

Where, $k$ and $b$ are constants in the derivation [45~47], and they can be expanded as follows

$$F_d = \frac{1}{2} C_d \cdot A \cdot \rho_a \cdot v^2 \tag{9}$$

$$F_r = (m_v + m_c) \cdot g \cdot \cos\theta \cdot f_R \tag{10}$$

$$F_g = (m_v + m_c) \cdot g \cdot \sin\theta \tag{11}$$

$$F_a = (\delta_i m_v + m_c) a_x \tag{12}$$

$$T_{Dem} = F_{Dem} * r_d = (F_a + F_G + F_R + F_D) * r_d$$
$$= \left[(\delta_i m_v + m_c) a_x + (\sin\theta + f_R \cdot \cos\theta)(m_v + m_c) g + C_D A \frac{\rho_a}{2} v^2\right] * r_d = kv^2 + b \tag{13}$$

$$k = C_D A \frac{\rho_a}{2} * r_d \tag{14}$$

$$b = [(\delta_i m_v + m_c) a_x + (\sin\theta + f_R \cdot \cos\theta)(m_v + m_c) g] * r_d \tag{15}$$

Where, $\delta_i$ is the vehicle rotating mass conversion factor [46], $m_v$ is the mass of the vehicle, $m_c$ is the mass of the cargo, $a_x$ is the acceleration of the vehicle, $\theta$ is the slope angle, $f_R$ is the road friction coefficient, $C_D$ is the drag coefficient, $A$ is the frontal area and $\rho_a$ is the air density [47]. Constraints are related to the maximum electric motor rotating speed, torque and the driving conditions. The electric motor rotating speed is lower than 1,600 rpm based on the data as shown in Figure 3. The maximum driving torque that the electric motor can output is 1,250 N·m. The working mode is assumed to be constrained by the motoring efficiency map as shown in Figure 3 (c). IWM-AEV can be optimized to find out the lowest energy consumption point under the specific upslope angle and initial speed using the demanding energy factor.

*3.4 Downslope driving condition*

The downslope energy analysis is slightly more complex than the upslope one since braking, coasting down and energy regenerating need to be taken into consideration. When IWMs are motoring, the calculation is similar to the upslope circumstance. If the electric motor works in a braking status, the motor works as a generator. The energy regeneration in the electric vehicle's powertrain system from wheels to the battery can be modeled through [44]

$$E_{reg} = E_{recover} - E_{SSCM} = P_{recover}t - P_{SSCM}t = \frac{T_{dem,2}*n*\eta_b*\eta_c*\eta_{recover}*\eta_t*\eta_i}{9550} * \frac{1}{v} - P_{SSCM}t$$

$$= \phi\eta_r \int_{a>0} F_a v dt - \frac{1}{\eta_d \eta_i \eta_m \eta_t}\left(\int F_d v dt + \int F_r v dt + \int F_g v dt + \int_{a>0} F_a v dt\right) - P_{SSCM}t \quad (16)$$

Where, $E_{reg}$ (Wh) is the regenerated energy per km, $P_{recover}$ (kW) is the energy recovery of the traction system, $\eta_b$ is the IWM braking efficiency, $\eta_c$ is the battery charging efficiency, $\eta_{recover}$ is the IWM braking energy recovery rate, $T_{dem,2}$ (N·m) is the torque demand to balance the electric vehicle when it is braking, $n$ (rpm) is the rotating speed and $v$ (km h$^{-1}$) is the driving speed. Defining a parameter called regenerating energy factor ($\lambda_2 = T_{dem,2} * \eta_b$) and using the relationship of rotating speed and driving speed as shown in Eq. (6), the unit regenerated energy is a linear function of the regenerating energy factor:

$$P_{recover} = \frac{30*\eta_c*\eta_{recover}*\eta_t*\eta_i}{\pi*r_d*3.6*9550} * \lambda_2 \quad (17)$$

Where, the value of the parameters used in the Eq. (17) is from Table 2. It can be seen that the regenerated energy per km of the autonomous electric vehicle under the downslope driving condition is linearly correlated to the regenerating energy factor, which needs to be further analyzed to examine the relationship among regenerated energy, slope angle and speed.

Under the downslope driving condition, the slope angles, the length of the slope, and the initial speed are given. When the AEV is motoring, the analysis is similar to the upslope circumstance. When the vehicle is braking, the decision variable is the vehicle driving speed $v$. The objective function is the regenerating energy factor ($\lambda_2$), which yields [45~47]:

$$\lambda_2 = T_{dem} * \eta_b = (kv^2 + b) * \eta_b \tag{18}$$

Where, $k$ and $b$ are given in Eqs. (14) and (15), respectively. The IWM-AEV can drive at the highest energy regeneration points under real-time downslope angles and speeds based on the regenerating energy factor we derived.

## 4. Results and discussion

### 4.1 Energy analysis of the flat road driving

In this study, the energy consumptions of the IWM-AEV on the flat road are analyzed under both UDDS (Urban Dynamometer Driving Schedule) and HWFET (Highway Fuel Economy Test) driving cycles, to reflect the different energy consumption resulting from different driving patterns. Based on Eq. (5), the unit energy consumption for each driving cycle ($E_{unit}$) can be simulated as [31].

$$E_{traction} = \frac{1}{\eta_d \eta_i \eta_m \eta_t} \left( \int F_d v dt + \int F_r v dt + \int F_g v dt + \int_{a>0} F_a v dt \right) - \phi \eta_r \int_{a>0} F_a v dt,$$

$$E_{unit} = \frac{E}{D_{cycle}} = \frac{E_{traction} + E_{SSCM}}{D_{cycle}} = E_{unit,traction} + E_{unit,SSCM} \tag{19}$$

Where, $D_{cycle}$ is the driving distance of each driving cycle.

Coupling IWM efficiency maps and data points in different driving cycles, unit energy consumption for both UDDS and HWFET driving are calculated. The corresponding IWM-AEV unit energy consumptions of UDDS and HWFET under flat road driving are 140.3 Wh km$^{-1}$ and 163.4 Wh km$^{-1}$, respectively, both around 5% lower than that of the baseline EV which are at 147.8 Wh km$^{-1}$ and 172.6 Wh km$^{-1}$, respectively, for the UDDS and HWFET driving. For the UDDS, the traction system and the SSCM consume 94.6% and 5.4% of the total energy. For the HWFET, the traction system and the SSCM consume 98.1% and 1.9% of the total energy. The energy consumption of UDDS driving is lower than that of HWFET is because of the braking energy regeneration during local driving. The HWFET driving cycle has a higher average speed, which causes a low SSCM energy consumption on a unit driving distance basis. Although the SSCM system consumes some energy in monitoring and detecting the road conditions, the IWM-AEV still has better energy performance than the baseline EV due to its higher energy transmitting efficiency and lower vehicle mass. This result also justifies the advantages of the IWM layout from the energy-saving aspect.

*4.2 Energy analysis of the upslope driving*

Since there are no representative driving cycles for upslope driving, our analysis focuses on energy consumption at three typical upslope angles. The energy analysis on 5 degrees, 10 degrees and 15 degrees with a speed range between 0~120 km h$^{-1}$ was carried out. Under the 5, 10 and 15 degree upslope driving, the upslope speed of the lowest energy consumption and the energy consumptions of the IWM-AEV are 36 km h$^{-1}$ and 538.7 Wh km$^{-1}$, 52 km h$^{-1}$ and 981.7

Wh km$^{-1}$, 64 km h$^{-1}$ and 1435.5 Wh km$^{-1}$, respectively. Figure 4 shows the relationship between energy consumption and driving speed for the IWM AEV under different upslope angles. It can be seen that under the upslope driving, a larger slope angle leads to higher energy consumption. For each slope angle, the energy consumption declines first and then rises as the vehicle speed increases. The bounce trend results from the characteristics of the IWM, a type of direct current brushless electric motors which have a low motoring efficiency when the working torque or speed is close to the boundaries of the efficiency map [48]. The SSCM system will also consume more energy at a low speed due to a longer operating time per unit distance driven. At a high vehicle speed, although the SSCM system consumes less energy, the energy used to overcome the aerodynamic resistance is large. As a result, the AEV is simulated in our analysis under a constant upslope speed at the lowest energy consumption, as described above.

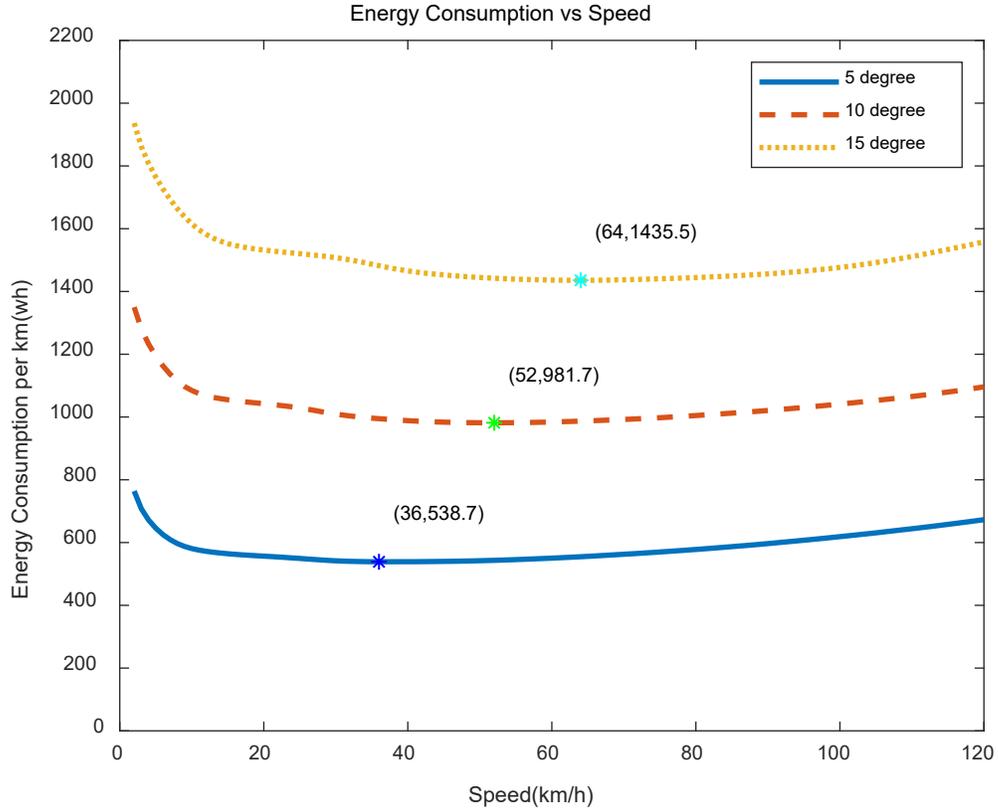

Figure 4: Energy consumption of the IWM-AEV at 0~120 km h$^{-1}$ speed under the upslope driving.

*4.3 Energy analysis for the downslope driving*

There could be two states during the downslope driving: braking state and motoring state. The working state is determined by the relationship between the slope resistance ($F_g$) and the sum of rolling resistance and aerodynamic resistance ($F_r+F_d$). In the braking state ($F_g > (F_r+F_d)$), the IWM-AEV drives down along the road and the IWMs will work as generators to recover energy. Under the motoring state ($F_g <(F_r+F_d)$), the slope is not steep enough and the IWM-AEV needs traction forces, which will consume energy. Following another IWM-EV study in literature [49], the initial speed is set at 30 km h$^{-1}$ to calculate the aerodynamic resistance ($F_d$) under the downslope driving case. The IWM-AEV in the downslope driving is also analyzed for various speeds ranging between 0-120 km/h.

*4.3.1 Braking state*

The IWM-AEV will regenerate energy from the braking during the braking state (for this case, $F_g > (F_r+F_d)$). With the initial speed set at 30 km h$^{-1}$ [49], the energy consumption of the IWM-AEV driven on different slope angles (-5 degrees, -10 degrees and -15 degrees) at a speed ranging between 0~120 km h$^{-1}$ is analyzed. For the -5 degrees downslope driving, the downslope speed of the highest regenerating energy is calculated to be 40 km h$^{-1}$ and the IWM-AEV will regenerate 161.2 Wh km$^{-1}$ at 40 km h$^{-1}$. For the -10 degrees downslope driving, the downslope speed of the highest regenerating energy is calculated to be 54 km h$^{-1}$ and the IWM-AEV will regenerate 379.6 Wh km$^{-1}$ at 54 km h$^{-1}$. For the -15 degrees downslope driving, the downslope speed of the highest regenerating energy is calculated to be 68 km h$^{-1}$ and the IWM-AEV will regenerate 585.9 Wh km$^{-1}$ at 68 km h$^{-1}$.

Figure 5 shows the relationship between the regenerated energy and slope speed under different downslope angles. Overall, the regenerated energy increases as the slope angle increases for the same speed and the speed corresponding to the highest regenerated energy gets larger as the slope angle increases. The regenerating energy factor is directly affected by the slope angle, and it controls the main part of the regenerated energy. For each slope angle, the speed of the highest regenerated energy gets larger when the slope angle becomes larger. The regenerated energy will first increase and then drop as the vehicle speed increases since the IWM braking efficiency is low when the motor works under extreme working conditions. When the vehicle speed is too low, the regenerated energy is not capable to overcome the energy needed for SSCM systems and this explains why the curves do not start at the axis origin. When the speed is high, the aerodynamic resistance consumes more energy to overcome, resulting in less regenerated energy.

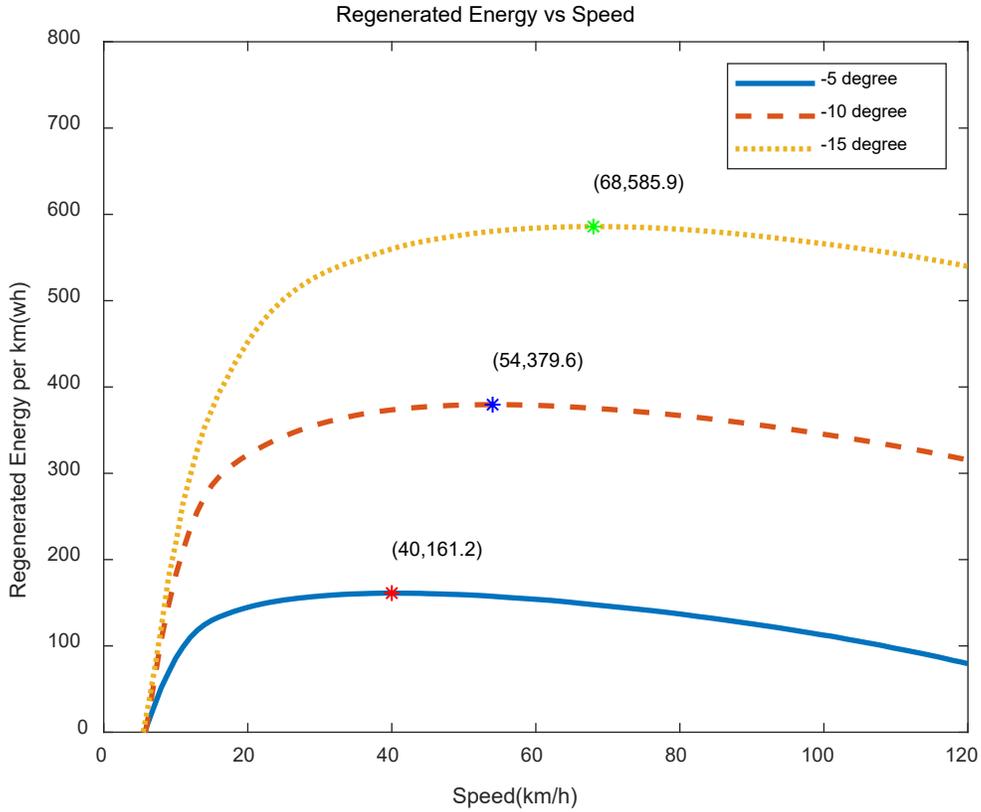

Figure 5: Regenerated energy of the IWM-AEV at 0~120 km h$^{-1}$ speed under the downslope braking.

*4.3.2 Motoring state*

The IWM-AEV will work under the motoring state and consume energy to power the vehicle when the slope angle is small. Setting the initial speed at 30 km h$^{-1}$ [49], the energy analysis on different slope angles (-0.2, -0.5 and -0.8 degrees) with a speed range between 0~120 km h$^{-1}$ was examined. For the -0.2 degrees downslope driving, the downslope speed of the lowest energy consumption is calculated to be 31 km h$^{-1}$ and the IWM-AEV will consume 63.9 Wh km$^{-1}$ at 31 km h$^{-1}$ using Eq. (16~18). For the -0.5 degrees downslope driving, the downslope speed of the lowest energy consumption is calculated to be 31 km h$^{-1}$ and the IWM-AEV will consume 49.2 Wh km$^{-1}$ at 31 km h$^{-1}$. For the -0.8 degrees downslope driving, the downslope speed of the lowest energy consumption is calculated to be 31 km h$^{-1}$ and the IWM-AEV will consume 14.6 Wh km$^{-1}$ at 31 km h$^{-1}$. Figure 7 shows the relationship between the energy consumption and driving speed under these downslope angles. When the slope angle is small and the IWM is working under the motoring state, the energy consumption curve is similar to that of the upslope condition and the energy consumption will be smaller than that under the flat road (0 degrees) driving condition. A larger slope angle will lead to a lower energy consumption in the downslope motoring state.

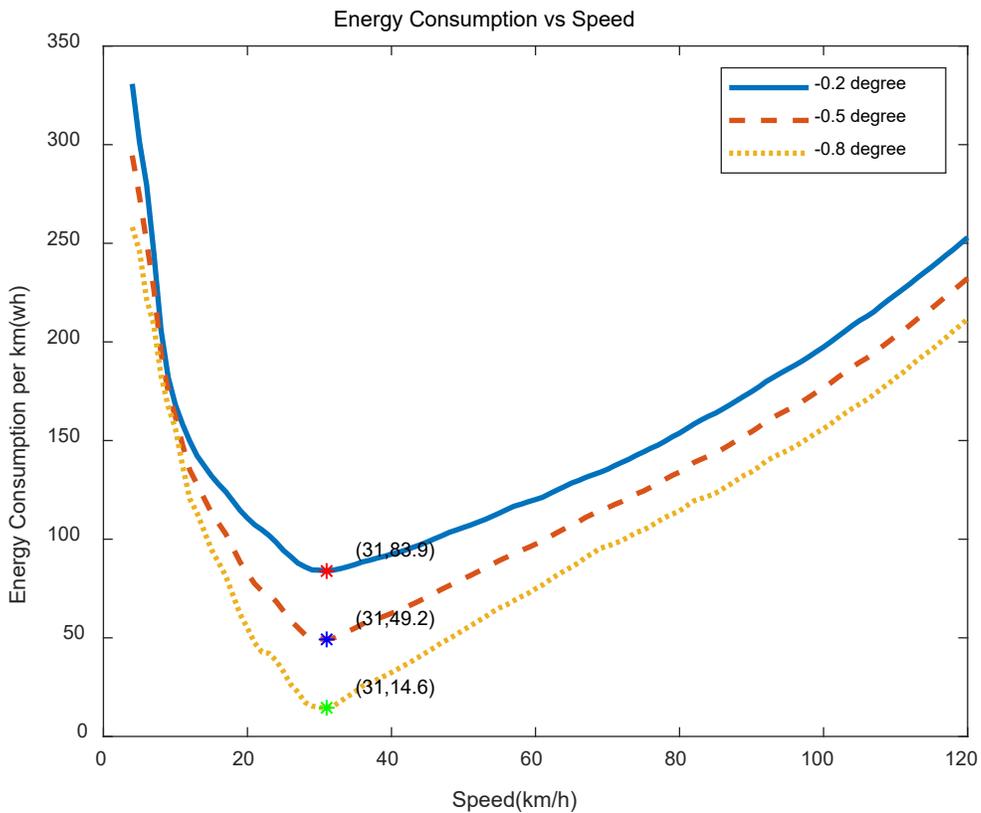

Figure 6: Energy consumption of the IWM-AEV at 0~120 km h$^{-1}$ speed under downslope motoring state.

*4.4 Validation*

The validation of the energy analysis model is conducted based on the energy consumption of a representative EV, i.e. Tesla Roadster since the IWM-AEV is still a conceptual technology under research and development. Compared with the energy consumption data of Tesla Roadster [50], the simulation result of the IWM-AEV on a flat road in our work has a similar trend (see Figure 7). The Tesla Roadster is 131 kg lighter than the IWM-AEV configured in this study, and its frontal area is 2.1 $m^2$, which is 23% smaller than the IMW-AEV. This is why the energy curve of Tesla Roadster is a little lower than that of the IWM-AEV. Both the IMW-AEV and the Tesla Roadster achieve their minimum energy consumption points at a speed of around 30 km h$^{-1}$. Thus, the energy analysis modeling and control strategies adopted in this study for the IWM-AEV are valid and acceptable.

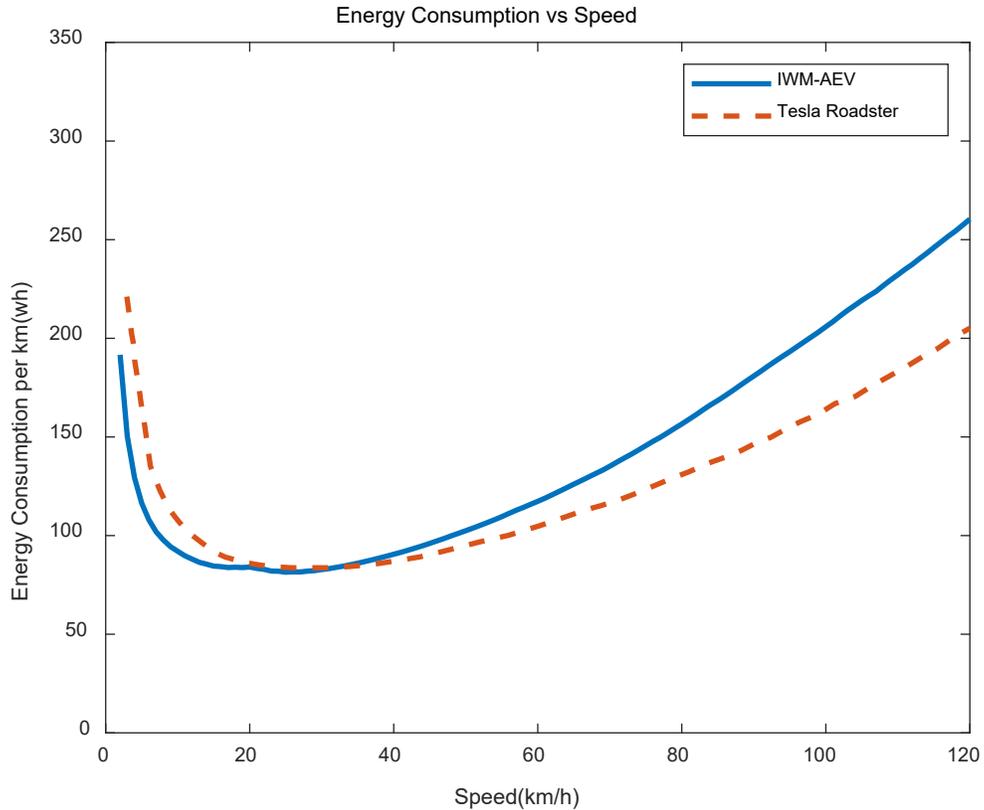

Figure 7: Comparison of energy consumption vs. speed for the IWM-AEV and Tesla Roadster under the flat road condition.

## 5. Case study

In this case study, the energy consumption of the baseline Nissan Leaf EV was analyzed to demonstrate the energy-saving potential of the configured IWM-AEV technology over conventional EV powertrain technology, based on an actual test dataset in a driver report [51]. In this case study, a comparison between the baseline EV and its IWM-AEV counterpart has been made under both upslope and downslope driving conditions.

According to this driver report, the baseline Nissan Leaf EV was driven on a round trip from West Los Angeles to Rosamond in the United States [51]. It was driven first upslope for 124.1 km and then downslope for 124.1 km with an average speed of 55 mph (88.5 km h$^{-1}$). The slope angle is around 0.30 degree and the average energy consumption for the whole trip was 157.9 Wh km$^{-1}$ on the baseline EV, as reported in [51]. In order to analyze the energy consumption of its IWM-AEV counterpart, the baseline Nissan Leaf is reconfigured to remove the energy consumption from the driver mass, auxiliary systems, and the acceleration resistance due to the assumed constant speed driving for the AEV. Based on the parameters in Table 2, the energy consumption of the baseline Nissan Leaf EV with a conventional powertrain is simulated using the vehicle dynamic equations (Eqs. 9~15). Following our previous analysis method in [36], considering the driver mass as an average American adult bodyweight at 89.7 kg [52], the energy consumption caused by the acceleration resistance, auxiliary systems and driver mass accounts for 17.3%, 2.7% and 6.9%, respectively, of the energy consumption of the baseline Nissan Leaf EV. After the three energy consumers are removed, the recalculated average energy consumption is 116.8 Wh km$^{-1}$ for the whole trip of the baseline Nissan Leaf EV, which can then be used to compare with the energy consumption of the IWM-AEV to benchmark the energy efficiency of the conventional powertrain and IWM drivetrain technologies.

The round trip is simulated in two parts for the IWM-AEV. First, the energy consumption of the IWM-AEV under upslope driving at a 0.3-degree upslope angle is modeled. Using Eqs. (5~8), the upslope speed of the lowest energy consumption is calculated to be 25 km h$^{-1}$ (see Figure 8) and under this constant speed driving the IWM-AEV will consume 109.9 Wh km$^{-1}$. Second, the energy consumption of the IWM-AEV under downslope driving at a -0.3-degree downslope angle is modeled. To reflect the continuous driving, the optimal speed with the lowest energy consumption in the upslope simulation (25 km h$^{-1}$) is used as the initial speed for the downslope driving. Calculated by the Eqs. (16~18), the simulation result shows that the IWM-AEV needs motoring and the downslope speed corresponding to the minimum energy consumption is 31 km h$^{-1}$. It will cost the IWM-AEV 82.9 Wh km$^{-1}$ during the downslope driving. As a result, the average energy consumption of the IWM-AEV during the whole travel is 96.4 Wh km$^{-1}$. From the results, it shows that the energy consumption of the IWM powertrain is 17.5% lower than that of the conventional powertrain.

## Conclusions

Considering vehicle dynamics and autonomous driving pattern, a numerical energy model was developed for the in-wheel motor driven autonomous electric vehicle in this study. A bottom-up analysis was conducted to quantify the energy consumption and energy-saving potential under three driving conditions of flat road, upslope and downslope. The analysis results show that the in-wheel motor driven autonomous electric vehicle consumes 140.3 Wh km$^{-1}$ in Urban Dynamometer Driving Schedule and 163.4 Wh km$^{-1}$ during Highway Fuel Economy Test driving cycles, respectively, on the flat road driving. Simulation results were further compared to a baseline electric vehicle driving data in West Los Angeles in the case study, and it was found that an in-wheel motor driven autonomous electric vehicle can potentially save 17.5% of energy over a conventional electric vehicle during slope driving. The in-wheel motor driven autonomous electric vehicle energy analysis is in line with actual working situations of commercialized electric vehicles and in-wheel motors, and thus it will be useful in supporting the development and applications of future sustainable autonoumous electric vehicle designs. In the future, we will work on the control algorithm to implement the energy-saving method proposed in the work. A test bench will also be built to further validate the simulation results.

**References**

[1] Vražić M, Barić O, Virtič P. Auxiliary systems consumption in electric vehicle. Przegląd elektrotechniczny. 2014 Jan 1;90(12):172-5.

[2] Yang F, Xie Y, Deng Y, Yuan C. Predictive modeling of battery degradation and greenhouse gas emissions from US state-level electric vehicle operation. Nature communications 2018;9(1):2429.

[3] Hadian M, AlTuwaiyan T, Liang X, Zhu H. Privacy-preserving Task Scheduling for Time-sharing Services of Autonomous Vehicles. IEEE Trans Veh Technol 2019;68(6):5260-70.

[4] Al-Jazaeri AO, Samaranayake L, Longo S, Auger DJ. Fuzzy logic control for energy saving in autonomous electric vehicles. In2014 IEEE International Electric Vehicle Conference (IEVC) 2014 Dec 17 (pp. 1-6). IEEE.

[5] Zhang TZ, Chen TD. Smart charging management for shared autonomous electric vehicle fleets: A Puget Sound case study. Transportation Research Part D: Transport and Environment. 2020 Jan 1;78:102184.

[6] De Santiago J, Bernhoff H, Ekergård B, Eriksson S, Ferhatovic S, Waters R, Leijon M. Electrical motor drivelines in commercial all-electric vehicles: A review. IEEE Trans Veh Technol 2011; 61(2):475-84.

[7] Chau KT. Energy Systems for Electric and Hybrid Vehicles. The Institution of Engineering and Technology (IET); 2016.

[8] Zhuang W, Qu L, Xu S, Li B, Chen N, Yin G. Integrated energy-oriented cruising control of electric vehicle on highway with varying slopes considering battery aging. Science China Technological Sciences. 2020 Jan;63(1):155-65.

[9] Fernández-Rodríguez A, Fernández-Cardador A, Cucala AP, Falvo MC. Energy Efficiency and Integration of Urban Electrical Transport Systems: EVs and Metro-Trains of Two Real European Lines. Energies. 2019 Jan;12(3):366.

[10] Ristiana R, Rohman AS, Rijanto E, Purwadi A, Hidayat E, Machbub C. Designing optimal speed control with observer using integrated battery-electric vehicle (IBEV) model for energy efficiency. Journal of Mechatronics, Electrical Power, and Vehicular Technology. 2018 Dec 30;9(2):89-100.

[11] Murata S. Innovation by in-wheel-motor drive unit. Vehicle System Dynamics, 2012; 50(6):807-30.

[12] Huang X, Wang J. Model predictive regenerative braking control for lightweight electric vehicles with in-wheel motors. Proc Inst Mech Eng, Part D: J Automob Eng 2012;226(9):1220-32.

[13] Zhai L, Hou R, Sun T, Kavuma S. Continuous steering stability control based on an energy-saving torque distribution algorithm for a four in-wheel-motor independent-drive electric vehicle. Energies 2018;11(2):350.

[14] Wu DM, Li Y, Du CQ, Ding HT, Li Y, Yang XB, Lu XY. Fast velocity trajectory planning and control algorithm of intelligent 4WD electric vehicle for energy saving using time-based MPC. IET Intelligent Transport Systems 2018;13(1):153-9.

[15] Yone T, Fujimoto H. Proposal of a range extension control system with arbitrary steering for in-wheel motor electric vehicle with four wheel steering. 2014 IEEE 13th International Workshop on Advanced Motion Control (AMC). 2014.

[16] Jain M, Williamson SS. Suitability analysis of in-wheel motor direct drives for electric and hybrid electric vehicles. 2009 IEEE Electrical Power & Energy Conference (EPEC). 2019.

[17] Zhang H, Zhao W. Decoupling control of steering and driving system for in-wheel-motor-drive electric vehicle. Mechanical Systems and Signal Processing 2018; 101: 389-404.

[18] Gang L, Zhi Y. Energy saving control based on motor efficiency map for electric vehicles with four-wheel independently driven in-wheel motors. Advances in Mechanical Engineering 2018;10(8):1-18.

[19] Jiang X, Chen L, Xu X, Cai Y, Li Y, Wang W. Analysis and optimization of energy efficiency for an electric vehicle with four independent drive in-wheel motors. Advances in Mechanical Engineering. 2018 Mar;10(3):1687814018765549.

[20] Rimac Automobili. Rimac Automobili C_Two hypercar – A car alive with technology. https://ctwo.rimac-automobili.com [last accessed Jan 2021].

[21] Long G, Ding F, Zhang N, Zhang J, Qin A. Regenerative active suspension system with residual energy for in-wheel motor driven electric vehicle. Applied Energy. 2020 Feb 15;260:114180.

[22] Building Ford's Next-Generation Autonomous Development Vehicle; https://medium.com/@ford/building-fords-next-generation-autonomous-development-vehicle-82a6160a7965 [last accessed Jan 2021].

[23] VLP-16 Velodyne LiDAR Puck. USER'S MANUAL AND PROGRAMMING GUIDE. https://greenvalleyintl.com/wp-content/uploads/2019/02/Velodyne-LiDAR-VLP-16-User-Manual.pdf [last accessed Jan 2021].

[24] BOSCH. Chassis Systems Control Fourth generation long-range radar sensor (LRR4). http://cds.bosch.us/themes/bosch_cross/amc_pdfs/LRR4_292000P0ZH_EN_low.pdf [last accessed Jan 2021].

[25] Dragonfly 2 Technical Reference Manual. https://manualzz.com/doc/23765992/dragonfly-2---point-grey-research [last accessed Jan 2021].

[26] Ultrasonic Sensor (Gen. 4) for parking aid systems. https://hexagondownloads.blob.core.windows.net/public/AutonomouStuff/wp-content/uploads/2019/05/Neobotix_Ultrasonic_sensors_whitelabel.pdf [last accessed Jan 2021].

[27] PwrPak7 Installation and Operation User Manual, NovAtel. https://docs.novatel.com/oem7/Content/PDFs/PreviousVersions/PwrPak7_Installation_Operation_Manual_v1.pdf [last accessed Jan 2021].

[28] MK5OBU MK5 DSRC RADIO ROADSIDE UNIT User Manual Cohda Wireless Pty Ltd. https://fccid.io/2AEGPMK5OBU/User-Manual/User-Manual-2618971 [last accessed Jan 2021].

[29] NVIDIA. World's first functionally safe AI self-driving platform. https://www.nvidia.com/en-us/self-driving-cars/drive-platform [last accessed Jan 2021].

[30] Gawron, J. H., Keoleian, G. A., De Kleine, R. D., Wallington, T. J., & Kim, H. C. (2018). Life Cycle Assessment of Connected and Automated Vehicles: Sensing and Computing Subsystem and Vehicle Level Effects. Environmental science & technology, 52(5), 3249-3256.

[31] Zhang C, Yang F, Ke X, Liu Z, Yuan C. Predictive modeling of energy consumption and greenhouse gas emissions from autonomous electric vehicle operations. Applied Energy. 2019 Nov 15;254:113597.

[32] Hayes JG, Davis K. Simplified electric vehicle powertrain model for range and energy consumption based on epa coast-down parameters and test validation by argonne national lab data on the nissan leaf.  2014 IEEE Transportation Electrification Conference and Expo (ITEC). 2014.

[33] Nissan USA. Nissan Leaf 2011SL detailed data;2019. https://www.nissanusa.com/electric-cars/leaf  [last accessed January 2021].

[34] Dragica KP, Making the impossible, possible – overcoming the design challenges of in-wheel motors, 2012. https://www.proteanelectric.com/f/2018/04/MakingTheImpossiblePossible.pdf [last accessed Jan 2021].

[35] Alexander F, In-wheel electric motors - the packaging and integration challenges, 2011. https://www.proteanelectric.com/f/2018/04/In_Wheel_Electric_Motors_AFraser_ProteanV4.pdf [last accessed Jan 2021].

[36] Zhang C, Shen K, Yang F, Yuan C. Multiphysics Modeling of Energy Intensity and Energy Efficiency of Electric Vehicle Operation. Procedia CIRP. 2019 Jan 1;80:322-7.

[37] Stutenberg K. Advanced Technology Vehicle Lab Benchmarking-Level 1. 2014 US DOE Vehicle Technologies Program Annual Merit Review and Peer Evaluation Meeting. 2014.

[38] He P, Khaligh A. Design of 1 kW bidirectional half-bridge CLLC converter for electric vehicle charging systems. 2016 IEEE International Conference on Power Electronics, Drives and Energy Systems (PEDES). 2016.

[39] Maia R, Silva M, Araújo R, Nunes U. Electrical vehicle modeling: A fuzzy logic model for regenerative braking. Expert Systems with Applications 2015;42(22):8504-19.

[40] Protean Electric. PD18 datasheet, 2018. https://www.proteanelectric.com/f/2018/05/Pd18-Datasheet-Master.pdf [last accessed Jan 2021].

[41] Deng Y, Li J, Li T, Gao X, Yuan C. Life cycle assessment of lithium sulfur battery for electric vehicles. Journal of Power Sources. 2017 Mar 1;343:284-95.

[42] Fred Lambert. Tesla is upgrading Model S/X with new, more efficient electric motors, Apr 2019. https://electrek.co/2019/04/05/tesla-model-s-new-electric-motors [last accessed Jan 2021].

[43] Li Z, Miotto A. Concentrated-winding fractional-slot synchronous surface PM motor design based on efficiency map for in-wheel application of electric vehicle. In2011 IEEE Vehicle Power and Propulsion Conference 2011 Sep 6 (pp. 1-8). IEEE.

[44] Chen Y, Xie B, Du Y, Mao E. Powertrain parameter matching and optimal design of dual-motor driven electric tractor. Int J Agric Biol Eng 2019;12(1):33-41.

[45] Rhode S, Hong S, Hedrick JK, Gauterin F. Vehicle tractive force prediction with robust and windup-stable Kalman filters. Control Engineering Practice 2016;46:37-50.

[46] Li L, Liu Q. Acceleration curve optimization for electric vehicle based on energy consumption and battery life. Energy 2019;169:1039-53.

[47] Ragatz A, Thornton M. Aerodynamic drag reduction technologies testing of heavy-duty vocational vehicles and a dry van trailer. 2016 [Technical Report].

[48] Van Niekerk D, Case M, Nicolae DV. Brushless direct current motor efficiency characterization. 2015 Intl Aegean Conference on Electrical Machines & Power Electronics (ACEMP), 2015 Intl Conference on Optimization of Electrical & Electronic Equipment (OPTIM) & 2015 Intl Symposium on Advanced Electromechanical Motion Systems (ELECTROMOTION). 2015.

[49] Zhao J, Ma Y, Zhao H, Cui Y, Chen H. PID slip control based on vertical suspension system for in-wheel-motored electric vehicles. In2018 Chinese Control And Decision Conference (CCDC) 2018 Jun 9 (pp. 1126-1131). IEEE.

[50] Straubel JB. Roadster efficiency and range. Tesla Motors. 2008 Dec 28;22. https://www.tesla.com/blog/roadster-efficiency-and-range [last accessed Jan 2021].

[51] Ben Stewart, Autoweek. Long-term 2018 Nissan Leaf: discovering the possibilities and limitations of the Leaf's EV Range, 2018. https://autoweek.com/article/car-reviews/long-term-2018-nissan-leaf-discovering-possibilities-and-limitations-leafs-ev [last accessed Jan 2021].

[52] Fryar CD, Kruszan-Moran D, Gu Q, Ogden CL. Mean body weight, weight, waist circumference, and body mass index among adults: United States, 1999–2000 through 2015–2016. National Health Statistics Reports, No. 122. Dec 2018;https://www.cdc.gov/nchs/data/nhsr/nhsr122-508.pdf [last accessed Jan 2021].